\documentclass{article}
\usepackage[nonatbib,preprint]{neurips_2020}

\usepackage[utf8]{inputenc} 
\usepackage[T1]{fontenc}    
\usepackage{hyperref}       
\usepackage{url}            
\usepackage{booktabs}       
\usepackage{amsfonts}       
\usepackage{nicefrac}       
\usepackage{microtype}      

\usepackage[ruled,vlined,linesnumbered]{algorithm2e}
\usepackage{xspace}
\usepackage{caption} 
\usepackage{booktabs}
\usepackage{colortbl}
\usepackage{mathtools}
\usepackage{amsmath}
\usepackage{amssymb}
\usepackage{amsthm}
\usepackage{amsfonts}
\usepackage{cleveref}
\usepackage{soul}

\Crefformat{section}{Section~#2#1#3}
\Crefformat{proposition}{Proposition~#2#1#3}
\Crefformat{equation}{Eq.\,#2#1#3} 

\usepackage{bm}
\usepackage{xspace}
\usepackage{enumitem}
\usepackage[dvipsnames]{xcolor}
\usepackage{subcaption}
\usepackage{caption}

\usepackage{stackengine}
\usepackage{multirow}
\usepackage{array}
\usepackage{wrapfig}
\usepackage[final]{pdfpages}

\usepackage[colorinlistoftodos, textwidth=30mm, shadow, textsize=small]{todonotes}
\definecolor{babyblue}{rgb}{0.54, 0.81, 0.94}
\definecolor{citrine}{rgb}{0.89, 0.82, 0.04}
\definecolor{misocolor}{rgb}{0.16,0.27,0.86}
\definecolor{jbcolor}{rgb}{0.9,0.4,0.2}
\definecolor{bernacolor}{rgb}{0.9608,0.4863,0.00}
\definecolor{carlcolor}{rgb}{0.0,0.9863,0.30}
\definecolor{grey}{rgb}{0.3, 0.3, 0.3}

\definecolor{graphicbackground}{rgb}{0.96,0.96,0.8}
\definecolor{rouge1}{RGB}{226,0,38}  
\definecolor{orange1}{RGB}{243,154,38}  
\definecolor{jaune}{RGB}{254,205,27}  
\definecolor{blanc}{RGB}{255,255,255} 
\definecolor{rouge2}{RGB}{230,68,57}  
\definecolor{orange2}{RGB}{236,117,40}  
\definecolor{taupe}{RGB}{134,113,127} 
\definecolor{gris}{RGB}{91,94,111} 
\definecolor{bleu1}{RGB}{38,109,131} 
\definecolor{bleu2}{RGB}{28,50,114} 
\definecolor{vert1}{RGB}{133,146,66} 
\definecolor{vert3}{RGB}{20,200,66} 
\definecolor{vert2}{RGB}{157,193,7} 
\definecolor{darkyellow}{RGB}{233,165,0}  
\definecolor{lightgray}{rgb}{0.9,0.9,0.9}
\definecolor{darkgray}{rgb}{0.6,0.6,0.6}
\definecolor{babyblue}{rgb}{0.54, 0.81, 0.94}
\definecolor{citrine}{rgb}{0.89, 0.82, 0.04}
\definecolor{misogreen}{rgb}{0.25,0.6,0.0}
\definecolor{PalePurp}{rgb}{0.66,0.57,0.66}
\definecolor{todocolor}{rgb}{0.66,0.99,0.99}
\definecolor{pearOne}{HTML}{2C3E50}
\definecolor{pearTwo}{HTML}{A9CF54}
\definecolor{pearTwoT}{HTML}{C2895B}
\definecolor{pearThree}{HTML}{E74C3C}
\colorlet{titleTh}{pearOne}
\colorlet{bull}{pearTwo}
\definecolor{pearcomp}{HTML}{B97E29}
\definecolor{pearFour}{HTML}{588F27}
\definecolor{pearFith}{HTML}{ECF0F1}
\definecolor{pearDark}{HTML}{2980B9}
\definecolor{pearDarker}{HTML}{1D2DEC}

\hypersetup{
	colorlinks,
	citecolor=pearDark,
	linkcolor=pearThree,
breaklinks=true,
	urlcolor=pearDarker}

\newcommand{\simclr}{\texttt{SimCLR}\xspace}

\newcommand{\SGD}{\normalfont\texttt{SGD}\xspace}

\newcommand{\pa}[1]{\left(#1\right)}

\let\originalleft\left
\let\originalright\right
\renewcommand{\left}{\mathopen{}\mathclose\bgroup\originalleft}
\renewcommand{\right}{\aftergroup\egroup\originalright}

\newcommand{\CommaBin}{\mathbin{\raisebox{0.5ex}{,}}}
\newcommand*{\eqdef}{\triangleq}

\renewcommand{\epsilon}{\varepsilon}

\newcommand{\nothere}[1]{}

\usepackage{xspace}

\newcommand{\algo}{\texttt{BYOL}\xspace}
\newcommand{\byol}{\algo}

\newcommand{\lars}{\texttt{LARS}\xspace}

\newcommand{\BatchNormalization}{batch normalization\xspace}
\newcommand{\WeightStandardization}{weight standardization\xspace}
\newcommand{\GroupNormalization}{group normalization\xspace}
\newcommand{\LayerNormalization}{layer normalization\xspace}
\newcommand{\InstanceNormalization}{instance normalization\xspace}
\newcommand{\batchnorm}{\texttt{BN}\xspace}
\newcommand{\layernorm}{\texttt{LN}\xspace}
\newcommand{\groupnorm}{\texttt{GN}\xspace}
\newcommand{\instancenorm}{\texttt{IN}\xspace}
\newcommand{\weightstandardization}{\texttt{WS}\xspace}

\newcommand{\netparams}{{\theta}}
\newcommand{\targetparams}{\xi}
\renewcommand{\eqdef}{\mathrel{\ensurestackMath{\stackon[1pt]{=}{\scriptscriptstyle\Delta}}}}
\renewcommand{\eqdef}{=}
\renewcommand{\gets}{=}
\newcommand{\goodinitperf}{65.7}
\newcommand{\wsperf}{73.9}

\title{BYOL works \textit{even} without batch statistics}

\author{%
Pierre H.\,Richemond\thanks{Equal contribution; the order of first authors was randomly selected.}\;$^{ ,1, 2}$ 
Jean-Bastien Grill\footnotemark[1]   $^{ , 1}$ \;
Florent Altché\footnotemark[1] $^{ , 1}$ \;
Corentin Tallec\footnotemark[1] $^{ , 1}$ \; 
Florian Strub\footnotemark[1] $^{ , 1}$ \; \AND
Andrew Brock$^{1}$\;
Samuel Smith$^{1}$\;
Soham De$^{1}$\;
Razvan Pascanu$^{1}$\; \AND
Bilal Piot$^{1}$ \;
Michal Valko$^{1}$\; \vspace{0.7em}\\
$^{1}$DeepMind \hspace{1cm} $^{2}$Imperial College\vspace{.7em}\\
\texttt{phr17@ic.ac.uk} \hspace{0.5cm} \texttt{[jbgrill,fstrub,altche,corentint]@google.com}
}

\begin{document}

\maketitle

\begin{abstract}
  Bootstrap Your Own Latent (\byol) is a self-supervised learning approach for
  image representation. From an augmented view of an image, \byol trains
  an online network to predict a target network representation of a different augmented view of the same 
  image. Unlike contrastive methods, \byol does not explicitly use a repulsion term built from 
  \textit{negative pairs}  in its training objective. Yet, it avoids
   collapse to a trivial, constant representation.
  Thus, it has recently been hypothesized that \BatchNormalization (\batchnorm) is critical 
  to prevent collapse in \byol. Indeed, \batchnorm flows gradients across
  batch elements, and could
  leak information about
  negative views in the batch, which could act as an implicit negative (contrastive) term. 
  However, we experimentally show that replacing
  \batchnorm with a batch-independent normalization scheme  (namely, a combination of group normalization and weight standardization)
  achieves performance comparable to vanilla \byol ($\wsperf \%$ vs. $74.3\%$ top-1 accuracy under the linear evaluation
  protocol on ImageNet with ResNet-$50$). Our
  finding disproves the hypothesis that the use of batch statistics is
  a crucial ingredient for \byol to learn useful
  representations.
\end{abstract}


\section{Introduction}

Self-supervised image representation methods~\cite{chen2020simple,SIMCLR2, MOCOv2, grill2020bootstrap} have achieved downstream performance that rivals those of supervised pre-training on ImageNet~\cite{ILSVRC15}. Current self-supervised methods rely on image transformations to generate different \emph{views} from an input image while preserving semantic information. 
Among the most successful algorithms, contrastive methods~\cite{CPC, CPCv2, he2019momentum, chen2020simple, MOCOv2, AMDIM}
use a loss function that balances out two terms: a term associated to the positive pairs (that we refer to as the \emph{positive} term) encouraging representations from views
of the same image to be similar, and a term associated to negative pairs (a \emph{negative} term) which encourages representations to be spread out. 


Taking a different route, other approaches manage to avoid the contrastive paradigm
~\cite{tarvainen2017mean,caron2018DeepCF,caron2020unsupervised,SpectralNet}.
Among them, \byol~\cite{grill2020bootstrap} learns
its representation by
predicting the target network representation of a view from the online representation of another view of the same image. However, such a setup has obvious \emph{collapsed} equilibria where the representation is constant, and thus can be predicted from any input.
This has raised the question of how \byol could even work without a negative 
term nor an explicit mechanism to prevent collapse.
Experimental reports~\cite{albrecht2020, tian2020understanding} suggest that the use of \BatchNormalization, \batchnorm~\cite{BatchNorm}, 
in \byol's network is crucial to achieve good performance. These reports hypothesise that the 
\batchnorm used in \byol's network could implicitly introduce a negative term. 

We experimentally confirm the particular importance of \batchnorm in \byol: removing all
instances of \batchnorm in the network prevents \byol from learning anything at all in the classic setting, see Section~\ref{sec:3.1} and Table~\ref{tab:normalization-ablation}. 
However, our experimental results given in Table~\ref{tab:summary_results} go against some interpretations proposed notably in~\cite{albrecht2020, tian2020understanding}. In particular, we \textbf{refute the following hypotheses}: 

\begin{itemize}[leftmargin=*]
\item (H1) \textit{\algo needs \batchnorm because \batchnorm provides an implicit negative term required to avoid collapse.} 
In Section~\ref{sec:3.2}, we show that \byol avoids collapse and achieves $\goodinitperf\%$ top-1 accuracy on ImageNet under the linear evaluation protocol~\cite{Zhang2016ColorfulIC} \textit{without} 
using any normalization during training,  
by using both a better
initialization scheme and retaining the additional
trainable parameters scaling and bias ($\gamma$ and $\beta$) introduced by \batchnorm.

Therefore, unlike (H1), we hypothesize 
that the main role of \batchnorm is to make the network more robust to cases when the initialization is scaled improperly. Indeed,
proper initialization
is critical for deep
nets~\cite{mishkin2016need, edge_chaos, xiao2018dynamical, fixup}
and \byol suffers from a bad initialization in two ways: 
(i) as for any deep network, it makes optimization difficult 
and (ii) \byol's target network outputs will be ill-conditioned,  
which initially provides poor targets for the online network.

\item (H2) \textit{\byol cannot achieve competitive performance without the implicit contrastive effect provided by batch statistics. }
In Section~\ref{ss:gnws}, we show that most of this performance gap---$\goodinitperf\%$ achieved without \batchnorm vs.\,$74.3\%$ 
achieved \emph{with} \batchnorm---can be bridged without using batch statistics. Specifically, if we replace \batchnorm 
with a combination of \GroupNormalization, \groupnorm~\cite{wu2018group}, and \WeightStandardization, \weightstandardization~\cite{qiao2019weight}, while keeping standard initializations, \byol achieves $\wsperf\%$ top-1 accuracy.
\end{itemize}


\section{Background}

In this section, we adopt the notation
of~\cite{grill2020bootstrap}. Recall that $x$ denotes an image
and $v \eqdef t(x)$ and
$v' \eqdef t'(x)$ are two views of $x$ obtained from two independent
transformations $t$ and $t'$ sampled from a distribution
$\mathcal{T}$. These views are used as input to an encoder network to
obtain representations $y_\theta \eqdef  f_\theta(v)$ and
$y'_\theta \eqdef  f_\theta(v')$; and projections
$z_\theta \eqdef  g_\theta(y_\theta)$ and $z'_\theta \eqdef  g_\theta(y'_\theta)$.
We continue by a brief recap of standard contrastive methods and \byol.

\paragraph{InfoNCE} Most contrastive methods use variants of the InfoNCE~\cite{CPC}
loss to train their representation, 
\vspace{-0.3em}
\begin{equation*}
    \text{InfoNCE}_{\netparams} \eqdef
    \underbrace{
      -\frac{
        \langle z_\netparams, z'_\netparams\rangle
      }{
        \tau\cdot\big\|z_\netparams\big\|\cdot\big\|z'_\netparams\big\|
      }
    }_{\text{positive term}} + 
    \underbrace{
          \log
      \sum_{i}\exp\frac{
        \langle z_\netparams, z^i_\netparams\rangle
      }{
        \tau\cdot\big\|z_\netparams\big\|\cdot\big\|z^i_\netparams\big\|
       }
    }_{\text{negative term}}\CommaBin
\end{equation*}
where the $z_\netparams^i$ are projections from views of all images (including $z'_{\netparams}$ but not $z_{\netparams}$),
and $\tau$ is a temperature parameter.
The first term of this loss (the \emph{positive} term)
encourages projections of views of the same image to become similar,
while the second term (the \emph{negative} term) makes projections
of views from different images more dissimilar.
Such a loss
has a strong theoretical underpinning: 
minimizing this loss is equivalent to maximizing a lower bound 
on the mutual information between the representation of two views~\cite{poole2019variational}
which is tight when the function approximator 
is sufficiently expressive.

\paragraph{BYOL}

\byol trains its representation using both an online network (parameterized by 
$\netparams$) and a target network (parameterized by $\targetparams$). 
As a part of the online network,
it further
defines a predictor network~$q_\netparams$ that is used to predict target
projections $z'_\targetparams$ using online projections $z_{\netparams}$ as inputs.
Accordingly, the parameters of the online projection are updated following
the gradients of the prediction loss
\begin{equation*}
    \text{BYOL}_{\netparams} \eqdef 
    -\frac{
      \langle q_\netparams(z_\netparams), z'_\targetparams\rangle
    }{
      \big\|q_\netparams(z_\netparams)\big\| \cdot \big\|z'_\targetparams\big\|
    }\cdot
\end{equation*}
In turn, the target network weights $\targetparams$ are updated as an exponential
moving average of the online network's weights, i.e. $
\targetparams \leftarrow (1 - \eta) \targetparams + \eta \netparams$,
with $\eta$ being a decay parameter.
As $q_\netparams(z_\netparams)$ is a function of $v$ and $z'_\targetparams$ is a
function of $v'$, \byol's loss can be seen as a measure of similarity between the
views $v$ and $v'$ and therefore resembles the positive term of the InfoNCE loss.

\paragraph{Group normalization (GN)} \groupnorm~\cite{wu2018group} is an activation normalization method, like \batchnorm~\cite{BatchNorm},  \LayerNormalization (\layernorm~\cite{LayerNorm}), and \InstanceNormalization (\instancenorm~\cite{ulyanov2016instance}).
For an activation tensor $X$ of dimensions $(N, H, W, C)$, \groupnorm first splits channels into $G$ equally-sized groups, then normalizes activations with the mean and standard deviation computed over disjoint slices of size $(1, H, W, C/G)$.
The number of groups $G$ thus trades off between normalization over all channels ($G=1$, equivalent to \layernorm), and normalization over a single one ($G=C$, equivalent to \instancenorm). Importantly, \groupnorm operates independently on each batch element and therefore \textit{it does not rely on batch statistics}.






\paragraph{Weight standardization (WS)} \weightstandardization normalizes the weights corresponding to each activation using weight statistics. Each row of the weight matrix $W$ is normalized to get a new weight matrix $\widehat{W}$ which is directly used in place of $W$ during training. 
Only the normalized weights $\widehat{W}$ are used to compute convolution outputs but 
the loss is differentiated with respect to non-normalized weights $W,$
\begin{equation*}
    \widehat{W}_{i,j} = \frac{W_{i,j} - \mu_i}{\sigma_i}\CommaBin
    \quad \text{with} \quad
    \mu_i = \frac{1}{\mathcal{I}}\sum_{j=1}^\mathcal{I} W_{i,j}
    \quad \text{and} \quad
    \sigma_i = \sqrt{\epsilon + \frac{1}{\mathcal{I}}\sum_{j=1}^\mathcal{I}\pa{W_{i,j} - \mu_i}^2},
\end{equation*}
where $\mathcal{I}$ is the input dimension (product of input channel dimension and kernel spatial dimension); we set $\epsilon = 10^{-4}$. Contrary to \batchnorm, \layernorm, and \groupnorm, \weightstandardization does not create additional trainable weights.


\section{Experimental results}
While many metrics can be used to evaluate  self-supervised representations, we focus on classification accuracy on ImageNet~\cite{ILSVRC15} under the standard linear evaluation protocol~\cite{Zhang2016ColorfulIC} with a ResNet-$50$ architecture, with the same setup as~\cite{CPC,he2019momentum,chen2020simple, grill2020bootstrap}. Unless otherwise specified, we follow the
training setup and hyperparameters described in~\cite{grill2020bootstrap} when training \byol.

\subsection{Removing \batchnorm causes collapse}
\label{sec:3.1}

In \Cref{tab:normalization-ablation}, we explore the impact of
using different normalization schemes
in \simclr and \byol, by
using either \batchnorm, \layernorm, or removing normalization
in each component of \byol and
\simclr, i.e., the encoder, the projector (for \simclr and \byol), and
the predictor (for \byol only). 
First, we observe that removing all instances of \batchnorm in \byol
leads to performance that is no better than random. Noticeably, this is
specific to \byol as \simclr still performs reasonably well in this
regime. Nevertheless, solely applying \batchnorm to the ResNet encoder
is enough for \byol to achieve high performance.\footnote{
Some of these observations differ from the ones initially reported
in~\cite{albrecht2020}. Specifically, the authors observed a collapse
when removing \batchnorm in \byol's predictor and projector. This difference
could be linked to the use of the \SGD optimizer instead of \lars~\cite{LARS}.}

From these observations, \cite{albrecht2020} hypothesizes that 
\batchnorm implicitly introduces a negative contrastive term, which
acts as a crucial component to stabilize training
(H1). This hypothesis may seem further supported by the performance difference
between \simclr and \byol when replacing \batchnorm (which uses batch statistics)
with \layernorm which does not. 

However, we observe that \batchnorm seems to be mainly useful in the ResNet encoder, for which standard initializations are known to lead to poor conditioning~\cite{bjorck2018understanding,yang2018a}. Also \byol might be even more affected by improper initialization as it creates its own targets. Rather than (H1), we therefore hypothesize that the main contribution of \batchnorm in \byol is to compensate for improper initialization.

\setlength\tabcolsep{4.6pt}
\begin{table}[!h]
   \small
    \centering
    \caption{ \textit{Ablation results on normalization, per network component:} The numbers correspond to top-1 linear accuracy (\%), 300 epochs on ImageNet, averaged over 3 seeds.}
    \begin{tabular}{c || c | c  | c | c | c | c | c | c | c | c | c | c | c | c | c }\toprule
        Encoder & \multicolumn{4}{c|}{\batchnorm} & \multicolumn{4}{c|}{\layernorm} & \multicolumn{4}{c|}{-} & \multicolumn{3}{c}{-} \\
        Projector & \multicolumn{2}{c|}{\batchnorm} & \multicolumn{2}{c|}{-} & \multicolumn{2}{c|}{\layernorm} & \multicolumn{2}{c|}{-} & \multicolumn{2}{c|}{\ \ \batchnorm} & \multicolumn{2}{c|}{\layernorm} & \multicolumn{3}{c}{-} \\
        Predictor & \batchnorm  & - & \batchnorm & - & \layernorm & - & \layernorm & - & \batchnorm & - & \layernorm & - & \batchnorm & \layernorm & - \\
        \hline
        \byol & $73.2$ & $73.2$ & $72.0$ & $72.1$ & $0.1$ & $5.4$ & $0.1$ & $0.1$ & $62.6$ & $0.1$ & $0.1$ & $0.1$ & $61.1$ & $0.1$ & $0.1$ \\
        \hline
        \simclr & \multicolumn{2}{c|}{$69.3$} & \multicolumn{2}{c|}{$68.5$} & \multicolumn{2}{c|}{$68.0$} & \multicolumn{2}{c|}{$67.8$} & \multicolumn{2}{c|}{$53.8$\footnotemark} & \multicolumn{2}{c|}{$56.7$} & \multicolumn{3}{c}{$0.1$} \\
        \bottomrule
    \end{tabular}
    \label{tab:normalization-ablation}
\end{table}
\footnotetext{Unstable in late training: three seeds ending at $48.4\%$, $57.9\%$, $56.1\%$.}

\subsection{Proper initialization allows working without \batchnorm} 
\label{sec:3.2}
To confirm this assumption, we design the following
protocol to mimic the effect of \batchnorm on
initial scalings and training dynamics, without using
or backpropagating through batch statistics.  Before training, we compute
per-activation
\batchnorm statistics for each layer by running a
single forward pass of the network with \batchnorm
on a batch of augmented data. 
We then remove then batch normalization layers,
but retain the scale and offset parameters $\gamma$ and $\beta$
trainable, and initialize them as
\begin{align*}
\gamma^k_\mathrm{init} \gets \frac{\gamma^k_0}{\sigma^k} \qquad \text{and} \qquad
\beta^k_\mathrm{init} \gets -\mu^k \cdot \gamma^k_\mathrm{init},
\end{align*}
where $\gamma^k_\mathrm{init}$ and $\beta^k_\mathrm{init}$ are the initialization of the
additional trainable parameters corresponding to
the $k$-th removed \batchnorm, and $\mu^k$ and
$\sigma^k$ are the batch statistics computed during
the first pass for the $k$-th removed \batchnorm.
Additionally, we set $\gamma^k_0 \gets 0$ if the $k$-th removed 
\batchnorm corresponds to the final \batchnorm layer
in a residual block, and $\gamma^k_0 \gets 1$ otherwise. This is similar
to what is done in~\cite{Goyal2017}, except that
we further rescale the initialization of the scale
and offset parameters by a data-dependent quantity, in the spirit of~\cite{krahenbuhl2015data}.
Such setup keeps the initial scaling effect of
\batchnorm, while avoiding the computation of
any batch statistic during training, thus discarding
any potential implicit contrastive term. 


We use the exact same hyperparameters as for
vanilla \byol (\emph{i.e.}, base learning rate of $0.2$, weight decay of $1.5 \cdot 10^{-6}$ 
and decay rate of $0.996$), except that we increase the number of warmup epochs from
$10$ to $50$. After $1000$ epochs, this representation
achieves $\goodinitperf$\% top-1 accuracy in the linear
evaluation setting compared to $74.3\%$ for the baseline. These results are reported in Table~\ref{tab:summary_results}.

Despite its comparatively low performance, the trained
representation still provides considerably better classification results
than a random ResNet-$50$ backbone, and is thus
necessarily not collapsed. This confirms that \byol
does not need \batchnorm to prevent collapse. It also
confirms that one of the effects of \batchnorm is to
provide better initial scalings and training dynamics,
and that, contrary to \simclr, these are required for \byol to
perform well. 

\begin{table}[!h]
    \centering
     \caption{\textit{Summary of our results}: top-1 accuracy with linear evaluation on ImageNet, at 1000 epochs.}
   
    \begin{tabular}{c c c c c} 
        \toprule
         \byol variant & Vanilla \batchnorm & No \batchnorm & Modified init. & \groupnorm + \weightstandardization \\ \midrule
         Uses batch statistics & Yes & No & No & No \\
         Top-1 accuracy (\%) & $74.3$ & $0.1$ & $\goodinitperf$ & $\wsperf$ \\  \bottomrule
    \end{tabular}
    \label{tab:summary_results}
\end{table}

\subsection{Using \groupnorm with \weightstandardization leads to competitive performance}
\label{ss:gnws}
In the previous section, we have shown that \byol can
learn a non-collapsed representation without using
\batchnorm. Yet, \byol performs worse in this
regime. This only disproves (H1), but \batchnorm
could still both provide better initial scaling
\emph{and} an implicit contrastive term, responsible
for some of the performance. To study this hypothesis, we explore other refined element-wise normalization 
procedures. More precisely, we apply weight standardization to convolutional and linear parameters by weight 
standardized alternatives, and replace all \batchnorm by \groupnorm layers.


To train the network, we use the same hyperparameters as
in \byol except for the weight decay, set to
$3\cdot 10^{-8}$ instead of $1.5\cdot 10^{-6}$, the
base learning rate set to $0.24$ 
instead of $0.2$ and the target update rate, set to $0.999$
instead of $0.996$; we also set the number of groups for \groupnorm to $G=16$. With this setup, \byol (+\groupnorm +\weightstandardization) achieves $\wsperf\%$ top-1 accuracy after 1000 epochs.

As neither \groupnorm nor \weightstandardization
compute batch statistics, this version of \byol
cannot compare elements from the batch, and
therefore it likewise cannot implement a batch-wise implicit
contrastive mechanism. Therefore, we experimentally show that \byol can maintain most of its performance even without a hypothetical implicit contrastive term provided by \batchnorm.
\section{Conclusion}


 
Unlike contrastive methods, the loss used in \byol does not explicitly
include a negative term that would encourage its representations to 
spread apart. 
Nonetheless, \byol's representation does not collapse during training, and
\batchnorm has been hypothesized to fill the crucial role of an implicit negative term
by leaking batch statistics into the gradient.
We refute this hypothesis, and show that \byol can achieve competitive results 
without using batch statistics. In particular,
\byol achieves $\goodinitperf\%$ top-1 accuracy when removing \batchnorm and changing the initialization. 
Moreover, \byol achieves a competitive $\wsperf\%$ top-1 accuracy by replacing \batchnorm with a normalization scheme operating element-wise.

\section*{Acknowledgement}
The authors would like to thank the following people for their help throughout the process of writing this paper, in alphabetical order: Jean-Baptiste Alayrac, Bernardo Avila Pires, Nathalie Beauguerlange, Elena Buchatskaya, Jeffrey De Fauw, Sander Dieleman, Carl Doersch, Mohammad Gheshlaghi Azar, Zhaohan Daniel Guo, Olivier Henaff, Koray Kavukcuoglu, Pauline Luc, Katrina McKinney, R\'emi Munos, Aaron van den Oord, Jason Ramapuram, Adria Recasens, Karen Simonyan, Oriol Vinyals and the DeepMind team. We would like to also thank the authors of the following papers for fruitful discussions: \cite{chen2020simple,AMDIM,albrecht2020,schwarzer2020data}.

\bibliographystyle{unsrt}
\bibliography{biblio}

\end{document}